\documentclass[conference]{IEEEtran}
\IEEEoverridecommandlockouts
\usepackage{cite}
\usepackage{amsmath,amssymb,amsfonts}
\usepackage{graphicx}
\usepackage{textcomp}
\usepackage{xcolor}
\usepackage{booktabs}
\usepackage{tabularx}
\usepackage{placeins}
\usepackage{tikz}
\usetikzlibrary{plotmarks}
\usepackage[hidelinks]{hyperref}
\graphicspath{{figures/}}
\providecommand{\Description}[1]{}

\ExplSyntaxOn
\cs_new_eq:NN \paper_original_paragraph \paragraph
\cs_new_protected:Npn \paper_paragraph_dispatch:nnn #1#2#3
  {
    \IfBooleanTF{#1}
      {\paper_original_paragraph*{#3}}
      {\IfNoValueTF{#2}
        {\paper_original_paragraph{#3}}
        {\paper_original_paragraph[#2]{#3}}}
  }
\RenewDocumentCommand \paragraph {s o m}
  {
    \tl_set:Nn \l_tmpa_tl {#3}
    \regex_replace_once:nnN {\.\s*\Z} {} \l_tmpa_tl
    \exp_args:NnnV \paper_paragraph_dispatch:nnn {#1}{#2} \l_tmpa_tl
  }
\ExplSyntaxOff
\renewcommand{\IEEEauthorrefmark}[1]{\textsuperscript{#1}}
\begin{document}
\title{Characterizing High Bandwidth Flash for LLM Serving}
\author{
\IEEEauthorblockN{Zack Yu\IEEEauthorrefmark{1}, Chloe Wong\IEEEauthorrefmark{1}, Coleman Hooper\IEEEauthorrefmark{1}, Minjae Lee\IEEEauthorrefmark{2}, Wonjun Kang\IEEEauthorrefmark{2},\\
Youngjin Cho\IEEEauthorrefmark{2}, Michael W. Mahoney\IEEEauthorrefmark{1}\IEEEauthorrefmark{3}\IEEEauthorrefmark{4}, Yakun Sophia Shao\IEEEauthorrefmark{1}, Kurt Keutzer\IEEEauthorrefmark{1}, Amir Gholami\IEEEauthorrefmark{1}\IEEEauthorrefmark{3}}
\IEEEauthorblockA{\IEEEauthorrefmark{1}University of California, Berkeley, Berkeley, California, USA\\
\IEEEauthorrefmark{2}FuriosaAI, Seoul, South Korea\quad
\IEEEauthorrefmark{3}ICSI, Berkeley, California, USA\quad
\IEEEauthorrefmark{4}LBNL, Berkeley, California, USA\\
\{zack.yu, chloe.wong, chooper, mmahoney, ysshao, keutzer, amirgh\}@berkeley.edu\\
\{minjae.lee, kangwj1995, youngjin.cho\}@furiosa.ai}
\thanks{Corresponding author: Amir Gholami (amirgh@berkeley.edu).}
}

\maketitle

\begin{abstract}
Large language model (LLM) serving requires substantial memory to store model weights and KV caches. Agentic workloads compound this pressure through repeated interactions over growing contexts, making it important to retain KV cache for reuse. High-bandwidth flash (HBF) offers a way to expand accelerator memory capacity for LLM serving, but its access costs and limited write endurance complicate its use. We evaluate HBF for high-throughput agentic serving across system design and scheduling choices to understand when additional capacity improves serving performance and energy efficiency. We introduce an HBM–HBF–host hierarchical storage system and buffered cache-aware scheduling, and use trace-driven simulations to analyze their effects on performance, energy consumption, and HBF write lifetime. Across the evaluated workloads, the fastest HBF-augmented systems reduce completion time by 21.1--86.9\% relative to HBM-only systems. Modeled energy savings reach 62.4\%, with benefits depending on the workload and weight placement. Buffered cache-aware scheduling extends estimated HBF write lifetime from 0.99 to 13.63 years in the evaluated configuration. These results demonstrate the importance of coordinating data placement and scheduling in serving LLM with HBF, positioning HBF as an actively utilized lower-level memory tier rather than passive storage.
\end{abstract}


\section{Introduction}
LLM serving places competing demands on accelerator memory \cite{gholami2024ai}. Model weights occupy substantial capacity, while KV cache grows with context length and the number of active requests. Larger batches improve weight reuse and kernel utilization but need more active KV cache~\cite{flexgen2023,vllm2023}, whereas retaining cached prefixes consumes capacity but avoids recomputation when later requests reuse them~\cite{sglang2024}. Agentic serving makes this balance particularly important. Repeated model calls can extend a shared context with new instructions, intermediate results, and tool outputs~\cite{autellix2025}. The KV cache of a completed request may therefore remain useful for a later call. Admitting more requests can displace the KV cache before it is reused, requiring host reloads or repeated prefill~\cite{continuum2025}.

High-bandwidth flash (HBF) offers a way to expand device memory and reduce dependence on host storage~\cite{sandiskhbf2025,h32026}. Its capacity can accommodate model weights, larger active KV caches, and more reusable prefixes. However, HBF has different costs from HBM. In the configuration studied here, HBF reads consume more energy than HBM reads, writes are much more expensive, and has limited write endurance~\cite{h32026,mapatterson2026}. Simply adding capacity does not ensure efficient serving.

We introduce an HBM--HBF--host hierarchical storage system that manages weight placement, KV residency, and cache movement across HBM, HBF, and host tiers. The more accessed data are placed on HBM, while less accessed data are placed on HBF to improve energy efficiency. We complement this hierarchical storage system with buffered cache-aware scheduling, which uses an admission buffer to reduce server pressure that evict reusable caches to improve write endurance.

We evaluate several possible HBF integration architectures with our hierarchical storage system and scheduling policy using a request-level simulator with kernel latency models fitted to GPU measurements. The simulator tracks scheduling, cache residency, data movement, and parallel execution while replaying multi-turn workloads for dense and sparse-MoE models. Our evaluation covers both the original traces and synthetically extended conversations, comparing completion time, modeled energy, cache reuse, and estimated write lifetime. 

This paper makes three contributions:
\begin{itemize}
\item An HBM--HBF--host memory hierarchy for high-throughput LLM serving. The added capacity lets the system keep more KV cache on the device. On the long-context GLM-5.2 trace, the daisy-chain design reduces completion time by 84.5\% and modeled energy by 62.4\% relative to HBM-only serving.
\item Buffered cache-aware scheduling that preserves reusable KV cache through admission control. For Qwen3-32B on the daisy-chain design, reserving 10\% admission headroom with cache-aware scheduling reduces completion time by 14.1\% and extends estimated HBF lifetime from 0.99 to 13.63 years relative to FCFS.
\item A trace-driven simulation analysis of how workload characteristics, weight  placement, and parallelism affect HBF’s value for high-throughput serving. The analysis identifies different sources of savings in dense and sparse serving and motivates HBF as an actively utilized lower-level memory tier rather than passive storage.
\end{itemize}

\section{Related Work}
\paragraph{LLM serving and KV-cache management.}
Prior work improves LLM serving throughput through batching~\cite{orca2022}, KV cache management~\cite{vllm2023,sglang2024}, and memory offloading ~\cite{flexgen2023,xie2026hisparse}. Agentic serving additionally introduce scheduling decisions to retain KV cache during tool calls for improved cache reuse ~\cite{autellix2025, continuum2025}. KV cache can also be retained beyond GPU memory using host storage or a distributed cache~\cite{lmcache2025,mooncake2025}. Our focus is coordinating these storage and scheduling decisions under HBF's asymmetric access costs and finite write endurance.

\paragraph{HBF and flash-based inference.}
Flash-based inference systems use flash to offload weights, swap KV cache, or perform computation~\cite{llminaflash2024,hifc2025,zhao2026llm}. Recent HBF proposals bring flash closer to the accelerator ~\cite{sandiskhbf2025,h32026}. Concurrent HBF serving studies highlight the challenges of storing KV caches in HBF, adopting SSD-style KV offloading or restricting HBF to static expert weights due to endurance problem~\cite{hbfsucks2026,petrucci2026hbf,son2026hbf}. We use scheduling to extend HBF lifetime and study how storing active KV cache in HBF improves serving efficiency.

\paragraph{Simulation-based evaluation.}
Recent simulators study heterogeneous serving infrastructure, disaggregated serving, and memory designs for agentic inference~\cite{llmservingsim2026,memexplorer2026}. We build an accurate request-level simulation to study how placement and admission decisions affect completion time, energy, and HBF writes in our proposed system.

\section{Methodology}
\label{sec:methodology}
\subsection{Goal of Study}
Our goal is to characterize how adding HBF improves LLM serving efficiency and what algorithms are necessary to obtain that improvement. HBF presents two main challenges, higher read energy than HBM and limited write endurance. We address these challenges with a hierarchical storage system and buffered cache-aware scheduling, respectively. We evaluate high-throughput serving mainly through workload completion time and modeled energy, and report mean time per output token (TPOT) to show the accompanying latency tradeoffs.

\subsection{Simulation Method}
\paragraph{Request execution and KV cache management.}
The simulator tracks request arrivals and simulates execution asynchronously across model replicas. Each step selects eligible work, performs required cache preparation, executes a forward pass, advances requests, and accounts for subsequent cache movement. Mixed batches can contain both prefill and decode work. Chunked prefill splits prefill across steps under a shared budget to protect latency \cite{agrawal2024sarathiserve}. Requests in a session execute in order, and the next turn becomes eligible after the preceding turn completes and its recorded gap elapses. The simulator tracks residency separately in HBM, HBF, host DRAM, and SSD. Prefixes already on the device avoid host reload, prefixes on the host incur transfer service, and unavailable prefixes are computed again. We model decode overflow and the resulting cache eviction and offload costs, including any reloads needed before subsequent execution. The simulator operates at the request level, tracking the number of tokens stored in each memory tier rather than individual token positions. We assume that token placement within each tier follows the specified policy.

\paragraph{Kernel Modeling.}

We calibrate kernel latency models using GPU measurements on NVIDIA H100 and B200. GEMM-style kernels use a roofline model. Grouped GEMM is calibrated using kernel profiles and accounts for routed tokens and active-expert weights. Attention is modeled separately for prefill, decode, mixed-batch, MLA, and sparse MLA~\cite{liu2025deepseek} using corresponding kernel profiles. Prefill latency mainly depend on number of query-key pairs, and decode latency mainly fits from memory bandwidth utilization. The indexer accounts separately for projections, scoring, top-k selection, and index writes. We model computation in FP16, except for indexer scoring which we model using FP8 measurements. HBF performance is extrapolated from the modeled kernels.

\paragraph{Energy Modeling.}
We model the dynamic energy of computation, memory accesses, and communication:
\begin{equation}
 E=\sum_j N_j e_j+\sum_m 8(R_m e_m^r+W_m e_m^w)
       +E_{\mathrm{comm}},
 \label{eq:energy}
\end{equation}
where $j$ indexes the arithmetic operations in Table~\ref{tab:arithmetic-energy}, $N_j$ is the operation count, and $e_j$ is the corresponding energy per operation. For memory tier $m$, $R_m$ and $W_m$ are bytes read and written, and $e_m^r$ and $e_m^w$ are read and write energy per bit. $E_{\mathrm{comm}}$ is the energy consumed by data transfers over communication links.

\paragraph{Parallelism Modeling.}
We model TP~\cite{megatron2019}, EP~\cite{gshard2021}, and data-parallel attention (DPA)~\cite{deepseekv32024}. With DPA enabled, each attention DP rank handles its own requests and KV cache, including full MLA QKVO projections and indexer work. FFN/MoE execution uses the combined rank batches and the configured TP/EP layout, with expert assignment balanced using LPLB~\cite{deepseek_lplb_2025}. The shared FFN phase waits for the slowest attention rank. For models using IndexCache~\cite{indexcache2026}, we separately model full layers, which execute the indexer, and shared layers, which reuse an earlier full layer’s sparse-token selection. For each layer category $g$, let $L_g$ denote the number of layers in that category. The forward latency is approximated as follows:
\begin{equation}
 T_{\mathrm{forward}}=\sum_g L_g
 \left(\max_r T^{\mathrm{attn}}_{g,r}
       +T^{\mathrm{FFN}}_g+T^{\mathrm{comm}}_g\right).
 \label{eq:dpa}
\end{equation}
Energy sums activity across ranks. Independent replicas maintain separate clocks and share modeled host resources. Total completion time is the latest replica completion.

\paragraph{Communication Modeling.}
We model concurrent transfers over independent PCIe links, accounting for bandwidth contention at shared host DRAM and SSD resources. Transfer rates are constrained by per-link bandwidth and the remaining available bandwidth of shared resources, and are updated dynamically as transfers start or complete. We approximate layer-wise KV prefetch from host storage~\cite{mooncake2025} with a prefetch depth of 2. Only the first two layers’ transfers are exposed and each remaining layers takes $\max(T_{\mathrm{layer}}, T_{\mathrm{reload}})$. GPU collective latency includes startup overhead, the maximum of network transfer time, and HBM buffer access time.

\begin{figure}[t]
 \centering
 \includegraphics[width=\columnwidth]{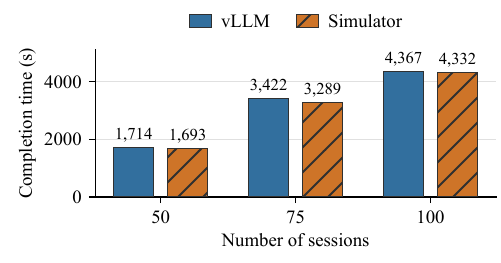}
 \caption{Measured vLLM and simulated completion time for Qwen3-235B-A22B on eight H100 NVL GPUs with eight-way tensor parallelism. Each pair replays the same multi-turn sessions using buffered cache-aware scheduling with 10\% admission headroom. Prefix reuse is confined to each session, and host KV offload is disabled.}
 \Description{Three pairs of bars compare vLLM and simulated completion time for 50, 75, and 100 sessions. Measured times are 1,714, 3,422, and 4,367 seconds. Simulated times are 1,693, 3,289, and 4,332 seconds.}
 \label{fig:vllm-comparison}
\end{figure}

\paragraph{Comparison with vLLM}
We compare our HBM-only simulator with vLLM~\cite{vllm2023} running Qwen3-235B-A22B~\cite{qwen32025} on eight H100 NVL GPUs with TP8. We replay subsets of 50, 75, and 100 multi-turn sessions from the LMCache trace, preserving their input lengths, output lengths, and inter-turn gaps. Both systems use buffered cache-aware scheduling with 10\% admission headroom, CUDA graphs, and per-session prefix caching. Hardware parameters like communication bandwidth and kernel utilization, are profiled locally. Figure~\ref{fig:vllm-comparison} compares workload completion time. Errors relative to vLLM are 1.21\%, 3.89\%, and 0.81\% (1.97\% on average).

\subsection{Simulation Trace}
The workload is derived from LMCache multi-turn agentic sessions~\cite{lmcache-agentic-dataset-2026}, containing 24,880 turns in 767 sessions. Input lengths are computed using each model's tokenizer, while output lengths and inter-turn gaps are taken from the dataset. We scale the workload by replicating conversations as independent sessions for the 100k-turn experiments. We also construct long trace (5$\times$ length) by concatenating copies of each session.

\paragraph{Sparse-attention lookup.}

We estimate selected-token placement from offline GLM sparse-selection profiles. At a fixed reference context length, we measure how many of a query’s selected tokens fall within the most recent positions, and interpolate between sampled window sizes to get this expected count for any HBM-resident window. At runtime, the number of selected tokens served from HBM is the profiled expectation, clamped to a feasible range. The remaining selected tokens are served from HBF, so the two tiers together always account for exactly the selected tokens.

\paragraph{Expert lookup.}

We profiled GLM-5.2’s expert routing on sessions from the LMCache trace. One subset serves as the training set, which determines expert-copy allocation and HBM placement, while the other subset serves as the test set, which estimates how many experts are accessed in each batch.


\section{System Overview}
\subsection{Exploring Future HBF Architectures}
\label{sec:architecture}
As HBF integration architectures are unsettled, we consider and evaluate three possible designs: daisy-chain, on-shoreline, and off-shoreline. The daisy-chain design is proposed in \cite{h32026}. It connects HBF behind HBM, retaining eight HBM stacks while adding eight HBF stacks, and maintain bandwidth through a latency-hiding buffer. The on-shoreline design is proposed in \cite{sandiskhbf2025}, where HBF stacks and HBM stacks compete for the limited shoreline space. The off-shoreline design is when HBF stacks are not placed on shoreline, but still near the accelerator. We study the daisy-chain design and on-shoreline design in the main results, and study the off-shoreline design in sensitivity analysis (Table~\ref{tab:sensitivity}) since its behavior would be similar to daisy-chain design with lesser bandwidth. For GEMM and decode attention, we model HBM and HBF accesses sequentially in the daisy-chain design and concurrently in the on-shoreline design. The daisy-chain design achieves equal or higher aggregate bandwidth by spreading accesses across more stacks, at the cost of more HBM/HBF stacks.


\subsection{Hierarchical Storage System}

\label{sec:hierarchy}

\begin{figure}[t]
 \centering\includegraphics[width=\columnwidth]{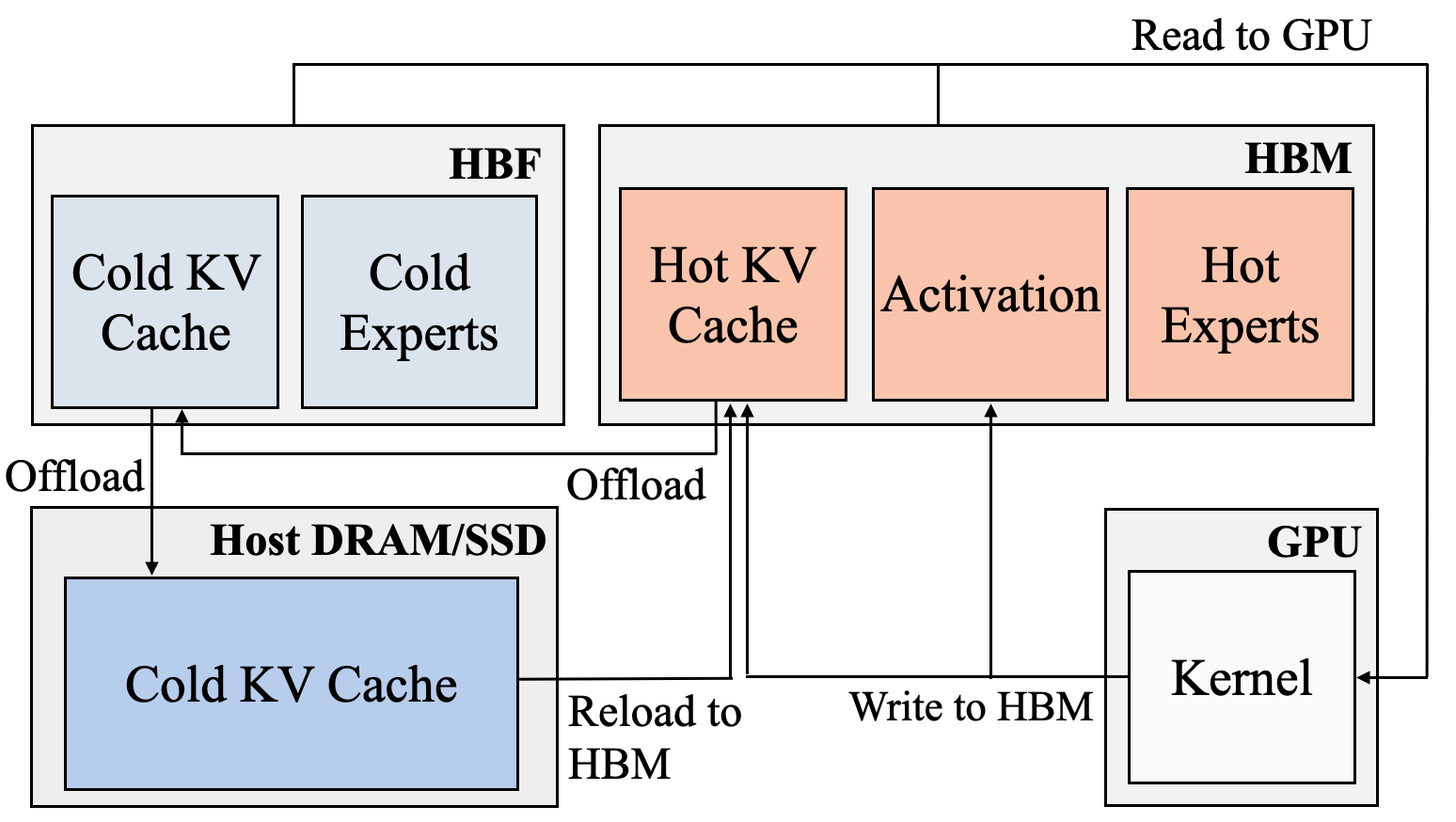}
 \caption{Hierarchical placement of weights, activations, and KV cache. Host reload returns to HBM, and colder state can move through HBF to the host.}
 \Description{A horizontal diagram shows host DRAM/SSD, HBF, HBM, and GPU compute. HBM holds hot KV, activations, and hot experts. HBF holds cold KV and experts. Arrows show reads to the GPU, writes to HBM, KV offload toward host storage, and host reload to HBM.}
 \label{fig:hierarchy}
\end{figure}

We propose a hierarchical storage system that leverages the sparsity of modern LLM architectures for energy efficiency. Figure~\ref{fig:hierarchy} shows the storage hierarchy. HBM contains frequently accessed weights and active data, while HBF contains the colder active data. Host DRAM and SSD hold data that are not actively in use. New KV cache and reloaded prefixes are staged in HBM. As space becomes scarce, older KV cache moves toward HBF and then the host.

\paragraph{Weights and activations.}
Weight placement is fixed for each evaluated configuration. Single-tier experiments place all weights in HBM or HBF, while the split MoE configuration places frequently used experts in HBM and the remainder in HBF. Activations are written to HBM since repeatedly writing short-lived activations to flash consume endurance rapidly.

\paragraph{Cache eviction and offload.}
We use request-level least-recently-used (LRU) ordering to select eligible cache entries when a memory tier runs out of space. Computed and reloaded KV cache writes to HBM first, and typically evict from HBM to HBF, then to host DRAM, and finally to SSD. Assuming LRU requests are less likely to be accessed than recently used requests, this retains hotter cache on HBM and moves colder cache to the cold memory tiers. Completed-session caches remain available for reuse until evicted. Reference locks exclude active requests from ordinary LRU eviction, although their KV cache can move from HBM to HBF when device capacity permits. 

\paragraph{Index-first placement.}
Sparse attention accesses only a selected subset of KV tokens, but the indexer accesses all index data across the context. We therefore treat index data as hotter data and separate the residency of indices and attention KV for more fine-grained data movement. Under HBM pressure, eligible KV is demoted before indices. On reload, indices preferentially return to HBM, displacing eligible KV when necessary. Under HBF pressure, indices are offloaded to the host before attention KV, so that indices can be placed in HBM on reload. These priorities aim to place the frequently accessed indices on HBM while placing more sparse KV in HBF.

\subsection{Buffered Cache-aware Scheduling}

\label{sec:scheduling}
We propose buffered cache-aware scheduling
to improve cache reuse for longer write lifetime. Cache-aware scheduling
prioritizes waiting requests with more device-resident prefix tokens and uses queue order to resolve ties. This avoids unnecessary reload or prefill work when a reusable prefix is available. However, if too many requests are admitted, their growing contexts can displace a completed session's KV cache before the next turn reuses it.


We add an admission fraction $b\in[0,1]$. Let $C$ be usable device KV/index capacity after resident weights, $L$ the currently locked footprint, and $A_i$ the footprint newly locked by admitting request $i$. Admission requires
\begin{equation}
 L+A_i\leq bC.
 \label{eq:buffer}
\end{equation}
A 10\% buffer uses $b=0.9$, while no buffer uses $b=1$. In the implementation, $A_i$ accounts for the incoming request's full context footprint, including any cached prefix newly brought under an active lock. Completed-session cache is evictable and is excluded from $L$ until re-admitted.

The threshold is only checked at admission. Active contexts can grow, and completed sessions can release locks while leaving cached data resident.  Although the buffer limits the active batch size, improved cache reuse can increase throughput. By retaining reusable KV cache, buffered scheduling reduces repeated writes to HBF and extends its estimated write lifetime as shown in Table~\ref{tab:scheduling}. The admission fraction $b$ also provides a way to adapt scheduling to device endurance and lifetime requirements. For devices with lower P/E endurance or deployments targeting a longer service life, a smaller $b$ leaves more headroom for cache retention and can further reduce HBF writes, at the cost of a smaller active batch.

\section{Results}
\label{sec:results}
\subsection{Experimental Setup}
\begin{table}[!t]
 \centering\small
 \caption{Per-stack memory assumptions. Given limited published HBF write specifications, we use typical SSD write parameters and assume SLC endurance of 100,000 P/E cycles.}
 \label{tab:memory}
 \begin{tabular}{lrr}
  \toprule
  Parameter & HBM & HBF\\
  \midrule
  Capacity (GB) & 24~\cite{micronhbm3e} & 375~\cite{h32026}\\
  Read bandwidth (GB/s) & 1,000~\cite{lenovob200} & 1,000~\cite{h32026}\\
  Write bandwidth (GB/s) & 1,000~\cite{lenovob200} & 8~\cite{zhao2026llm}\\
  Read startup ($\mu$s) & 0.1~\cite{mapatterson2026} & 20~\cite{h32026}\\
  Write startup ($\mu$s) & 0.1~\cite{mapatterson2026} & 250~\cite{hbfsucks2026}\\
  Read energy (pJ/bit) & 3.4~\cite{moon2023packaging} & 6.8~\cite{memexplorer2026}\\
  Write energy (pJ/bit) & 3.4~\cite{moon2023packaging} & 10~\cite{memexplorer2026}\\
  Assumed P/E cycles & --- & 100,000~\cite{son2026hbf}\\
  \bottomrule
 \end{tabular}
\end{table}

\begin{table}[!t]
 \centering\small
 \caption{Arithmetic energy parameters at 4~nm, scaled from 7~nm by 0.546~\cite{zhao2026llm, tsmcn52020,tsmcn4p2021}.}
 \label{tab:arithmetic-energy}
 \begin{tabular}{lr}
  \toprule
  Operation type $j$ & Energy $e_j$ (pJ/op)\\
  \midrule
  FP16 multiply--accumulate & 0.38 \\
  FP32 addition & 0.22 \\
  FP32 multiplication & 0.71 \\
  FP32 exponentiation & 7.4 \\
  FP32 division & 5.6 \\
  FP32 square root & 11.1 \\
  \bottomrule
 \end{tabular}
\end{table}

\begin{table}[!t]
 \centering\small
 \caption{Effects of scheduling algorithms on dense model serving. Qwen3-32B runs on one simulated B200 over 25k turns. $X+Y$ denotes HBM+HBF stacks per GPU. FCFS denotes first-come, first-served. Cache aware, 10\% reserves 10\% admission headroom. Life estimates HBF lifetime.}
 \label{tab:scheduling}
 \setlength{\tabcolsep}{2pt}
 \begin{tabular}{@{}llrrr@{}}
  \toprule
    \multicolumn{1}{@{}c}{Memory} &
  \multicolumn{1}{c}{Scheduling} &
  \multicolumn{1}{c}{\begin{tabular}[c]{@{}c@{}}Clock\\(s)\end{tabular}} &
  \multicolumn{1}{c}{\begin{tabular}[c]{@{}c@{}}Energy\\(MJ)\end{tabular}} &
  \multicolumn{1}{c@{}}{\begin{tabular}[c]{@{}c@{}}Life\\(years)\end{tabular}} \\
  \midrule
  8+0 & FCFS & 14,382.51 & 2.345 & ---\\
  8+0 & Cache aware & 13,910.18 & 2.306 & ---\\
  8+0 & Cache aware, 10\% & 12,138.70 & 2.289 & ---\\
  \midrule
  4+4 & FCFS & 15,889.41 & 2.948 & 0.51\\
  4+4 & Cache aware & 14,555.96 & 2.874 & 0.72\\
  \textbf{4+4} & \textbf{Cache aware, 10\%} & \textbf{12,852.15} & \textbf{2.804} & \textbf{8.25}\\
  \midrule
  8+8 & FCFS & 11,149.31 & 2.702 & 0.99\\
  8+8 & Cache aware & 9,755.60 & 2.668 & 4.23\\
  \textbf{8+8} & \textbf{Cache aware, 10\%} & \textbf{9,582.42} & \textbf{2.649} & \textbf{13.63}\\
  \bottomrule
 \end{tabular}
\end{table}

\begin{table*}[!t]
 \centering\small
 \setlength{\tabcolsep}{5pt}
 \renewcommand{\arraystretch}{1.0}
 \caption{Effects of memory architecture, weight placement, and parallelism on sparse-MoE serving. GLM-5.2 runs on 16 simulated B200s with a 10\% scheduling buffer. $X+Y$ denotes HBM+HBF stacks per GPU. HBM/HBF/Split indicates weight placement. Time denotes completion time and TPOT is mean per-request time per output token after the first. Bold marks minimum value of column.}
 \label{tab:glm-configurations}
 \begin{tabular}{@{}llllrrrrrr@{}}
  \toprule
  & & & & \multicolumn{3}{c}{1$\times$ trace} & \multicolumn{3}{c}{5$\times$ trace}\\
  \cmidrule(lr){5-7}\cmidrule(l){8-10}
  Architecture & Stacks & Parallelism & Weights & Time (s) & Energy (MJ) & TPOT (ms) & Time (s) & Energy (MJ) & TPOT (ms)\\
  \midrule
  HBM-only & 8+0 & TP16 & HBM & 7,959 & 7.08 & \textbf{161.13} & 151,620 & 76.79 & \textbf{151.01}\\
  \midrule
  Daisy-chain & 8+8 & TP16 & HBM & 3,853 & \textbf{5.47} & 219.51 & 23,508 & \textbf{28.89} & 282.23\\
  Daisy-chain & 8+8 & TP16 & HBF & 3,848 & 7.74 & 218.76 & 22,306 & 38.97 & 261.61\\
  Daisy-chain & 8+8 & $2\times$TP8 & HBF & \textbf{3,531} & 9.48 & 182.32 & \textbf{19,866} & 47.72 & 217.48\\
  Daisy-chain & 8+8 & $2\times$TP8 & Split & 3,700 & 6.99 & 197.09 & 21,144 & 36.75 & 244.45\\
  \midrule
  On-shoreline & 2+6 & TP16 & Split & 4,944 & 6.89 & 279.35 & 28,864 & 36.18 & 330.36\\
  On-shoreline & 4+4 & TP16 & Split & 4,408 & 6.30 & 258.46 & 27,450 & 33.22 & 337.10\\
  On-shoreline & 6+2 & TP16 & Split & 4,044 & 6.03 & 233.98 & 32,016 & 31.51 & 331.19\\
  \bottomrule
 \end{tabular}
\end{table*}

We study write endurance and scheduling using the dense Qwen3-32B model~\cite{qwen32025} on a single GPU, where small memory capacity would create higher pressure on KV cache management. We conduct a broader evaluation of sparse-MoE serving using GLM-5.2-FP16~\cite{glm2026} across different memory architectures, weight placement, and parallelism strategies. Table~\ref{tab:memory} gives per-stack parameters for HBM and HBF. The host has 16 DDR5-7200 channels (64~GB and 57.6~GB/s each) and 32 SSDs (1~TB and 8~GB/s each). We model PCIe Gen5 x16 at 63~GB/s per direction and NVLink at 900~GB/s per GPU endpoint.

Qwen weights reside in HBM except in on-shoreline 4+4, which splits them equally between HBM and HBF. GLM experiments use cache-aware scheduling with a 10\% buffer, mixed batches, session-sticky routing, DPA with one TP16/EP16 or two TP8/EP8 replicas, and the profiled lookup tables for sparse attention and MoE. Based on offline routing replay study, we use 32 redundant expert copies and assume ideal EP token balance under LPLB. Each scheduling engine shares a 2,048-token prefill budget across requests per step.

All sessions submit their first request at time zero. Subsequent turns arrive after the preceding turn completes and
the inter-turn gap elapses, so requests from different sessions
become staggered as execution progresses. Device and host hit rates are the fractions of input tokens found at admission in device memory and host storage. 

\subsection{Write Endurance and Dense Serving}

\paragraph{Scheduling and Write Endurance.}
Table~\ref{tab:scheduling} compares FCFS, cache-aware scheduling, and buffered cache-aware scheduling on Qwen3-32b. The configurations with HBF retain most reusable prefixes on the device with the buffer enabled. For daisy-chain, adding the buffer reduces HBF KV writes by 69.47\%, while completion time falls by 1.78\%. For on-shoreline 4+4, writes fall by 92.30\% and completion time by 11.71\%. We estimate HBF lifetime by dividing its total writable bytes (capacity times P/E endurance) by the simulated write rate (total bytes written to HBF over workload completion time). Assuming uniform wear, unit write amplification, and 100k P/E cycles, the reduced write rate raises estimated lifetime from 0.99 to 13.63 years for daisy-chain and from 0.51 to 8.25 years for on-shoreline 4+4 relative to FCFS. SanDisk’s announced 512 GB stacks are 16-high \cite{sandiskhbf2025}, implying 32 GB per die, well below the 128 GB capacity of mainstream TLC NAND dies\cite{samsung2024vnand}. SLC-grade endurance is therefore plausible.


\paragraph{Dense-model serving.}
With a 10\% scheduling buffer, daisy-chain reduces completion time by 21.06\% relative to HBM-only serving, while on-shoreline 4+4 increases it by 5.88\%. However, the corresponding energy costs increase by 15.76\% and 22.51\% respectively. This is mainly because the large amount of dense KV cache stored on HBF is more expensive to read, outweighing gains from weight reuse. 

\subsection{Memory Architecture, Weight Placement, and Parallelism}

We examine how memory architecture, weight placement, and parallelism affect serving efficiency for modern LLM architectures which notably leverages sparsity in data (Table~\ref{tab:glm-configurations}).

\paragraph{Architecture.}
We compare daisy-chain with on-shoreline designs across different HBM/HBF stack allocations. On the 1\(\times\) trace, daisy-chain with HBM weights reduces completion time and energy by 51.59\% and 22.76\% relative to HBM-only serving, while on-shoreline 6+2 achieves reductions of 49.19\% and 14.76\%. On the 5\(\times\) trace, these reductions increase to 84.50\% and 62.37\% for daisy-chain; 4+4 also becomes the fastest on-shoreline configuration, reducing completion time and energy by 81.90\% and 56.74\%. Longer contexts increase KV capacity needs, strengthening HBF's benefits and shifting the preferred on-shoreline allocation toward greater HBF capacity. Compared with HBM-only serving, the two architectures have 65.27\% higher TPOT on average. However, most agentic workloads are not latency-sensitive \cite{pan2026measuring}.

\paragraph{Weight placement.}
We compare HBM, HBF, and split weight placement within daisy-chain. On the 1$\times$ trace at TP16, moving weights from HBM to HBF changes completion time little but increases energy by 41.58\%. With two TP8 replicas, split placement reduces energy relative to all-HBF placement by 26.26\% and 22.98\% on the 1$\times$ and 5$\times$ traces, at completion-time costs of 4.80\% and 6.43\%. Moving weights to HBF frees HBM for KV cache but increases weight-read energy significantly. Retaining frequently accessed experts in HBM therefore balances capacity and energy efficiency.

\paragraph{Parallelism.}
We compare one TP16/EP16 replica with two TP8/EP8 replicas on 16 GPUs, keeping daisy-chain and all-HBF weight placement fixed. Two replicas reduce completion time by 8.26\% and 10.94\% on the 1$\times$ and 5$\times$ traces, while increasing energy by 22.51\% and 22.46\%. Mean TPOT also decreases on both traces. This configuration processes separate request groups with smaller communication groups, but requires larger per-GPU weight shards and duplicates model storage. Two replicas thus favor completion time and token latency, while one larger replica favors energy efficiency.

\subsection{Sources of Energy Savings and Its Implications}
Table~\ref{tab:decomposition} identifies weight reuse across larger batches as the main source of energy savings, especially for longer contexts. Qwen's dense attention reads the full KV cache, so the added cost of HBF reads outweighs the savings from weight reuse. GLM's sparse attention reads only selected KV entries, so the placement of frequently accessed KV entries on HBM can limit HBF read costs. These results suggest that HBF is most effective as an actively utilized lower memory tier, with frequently accessed data retained in HBM and less frequently accessed data served from HBF.

\begin{table}[!t]
 \centering\small
 \setlength{\tabcolsep}{3pt}
 \caption{Energy-savings breakdown relative to HBM-only serving. $X+Y$ denotes HBM+HBF stacks per GPU. Positive values indicates savings. Other savings are the residual.}
 \label{tab:decomposition}
 \begingroup
\small
\setlength{\tabcolsep}{3pt}
\renewcommand{\arraystretch}{1.0}
\begin{tabularx}{\columnwidth}{
  @{}>{\raggedright\arraybackslash}Xrrrrr@{}
}
\toprule
& \multicolumn{1}{c}{Qwen}
& \multicolumn{4}{c@{}}{GLM}\\
\cmidrule(lr){2-2}\cmidrule(l){3-6}
Contribution
& \multicolumn{1}{r}{\shortstack{8+8\\1$\times$}}
& \multicolumn{1}{r}{\shortstack{6+2\\1$\times$}}
& \multicolumn{1}{r}{\shortstack{8+8\\1$\times$}}
& \multicolumn{1}{r}{\shortstack{4+4\\5$\times$}}
& \multicolumn{1}{r@{}}{\shortstack{8+8\\5$\times$}}\\
\midrule
  Weight reads
  & $+19.86$ & $+12.60$ & $+20.52$ & $+52.71$ & $+57.46$\\
  KV/index reads
  & $-39.78$ & $-0.83$ & $-0.76$ & $-1.48$ & $-0.65$\\
  Other savings
  & $+4.16$ & $+3.00$ & $+3.00$ & $+5.51$ & $+5.57$\\
  \midrule
  \textbf{Net savings}
  & \textbf{-15.76\%}
  & \textbf{14.76\%}
  & \textbf{22.76\%}
  & \textbf{56.74\%}
  & \textbf{62.37\%}\\
  \bottomrule
\end{tabularx}
\endgroup
\end{table}

\begin{table}[t]
\caption{HBF parameter sensitivity for GLM-5.2 (daisy-chain, TP16, HBM weights).
Changes are relative to the Table~I baseline.  $\Delta$Energy denotes the change in energy savings.}
\label{tab:sensitivity}
\centering\small
\setlength{\tabcolsep}{3pt}
\begin{tabular}{@{}llrrrr@{}}
\toprule
 & & \multicolumn{2}{c}{1$\times$ trace}
   & \multicolumn{2}{c}{5$\times$ trace}\\
\cmidrule(lr){3-4}\cmidrule(lr){5-6}
Parameter & Value
& $\Delta$Time
& $\Delta$Energy
& $\Delta$Time
& $\Delta$Energy\\
\midrule
Read BW (GB/s)         & 500  & 0.00\% & 0.00\%    & +0.17\% & 0.00\% \\
                       & 250  & +1.10\% & +0.01\%   & +2.40\% & +0.06\% \\
Read energy (pJ/b)     & 10.2 & 0.00\% & $-$0.83\% & 0.00\%  & $-$0.79\% \\
                       & 13.6 & 0.00\% & $-$1.66\% & 0.00\%  & $-$1.58\% \\
Write energy (pJ/b)    & 20   & 0.00\% & $-$0.01\% & 0.00\%  & 0.00\% \\
Write startup ($\mu$s) & 350  & +0.24\% & +0.01\%   & +2.15\% & +0.04\% \\
Capacity (GB/stack)    & 512  & 0.00\% & 0.00\%    & 0.00\%  & 0.00\% \\
                       & 170  & 0.00\% & 0.00\%    & +5.83\% & +0.20\% \\
\bottomrule
\end{tabular}
\end{table}

\subsection{Sensitivity Analysis}
We vary HBF parameters for daisy-chain GLM-5.2 serving with TP16 and HBM weights, using Table~\ref{tab:memory} as the baseline. The read parameter sweeps explore the effects of reduced bandwidth in potential off-shoreline designs and uncertainty in production HBF read granularity.

\paragraph{Sensitivity to HBF reads.}
Table~\ref{tab:sensitivity} varies per-stack read bandwidth and energy per bit on both traces. Reducing bandwidth from 1,000 to 250~GB/s increases completion time by 1.1\% on the 1$\times$ trace and 2.4\% on the 5$\times$ trace. Mean TPOT rises by 5.33 and 9.04~ms respectively. At baseline bandwidth, doubling HBF read energy reduces the savings over HBM-only serving by 1.7\% on the 1$\times$ trace and 1.6\% on the 5$\times$ trace. Since HBF stores infrequently accessed KV cache, its performance and energy efficiency is less sensitive to degraded read parameters.

\paragraph{Sensitivity to HBF writes.}

Table~\ref{tab:sensitivity} varies write energy and write latency on both traces. Doubling write energy from 10 to 20~pJ/bit adds only +0.012\% energy cost. Increasing write startup latency from 250 to 350~$\mu$s increases completion time by 0.24\% on the 1 $\times$ trace and 2.15\% on the 5 $\times$ trace, and mean TPOT by 0.66 and 8.54~ms. This minimal difference is due to infrequent writes to HBF. Activations are written to HBM, model weights are read-only, and KV cache writes are significantly lesser than reads.

\begin{table}[t]
\caption{Stress test for HBF write lifetime under conservative cell-mode and endurance assumptions. Capacity is per stack. }
\label{tab:capacity-lifetime}
\centering\small
\setlength{\tabcolsep}{4pt}
\begin{tabular}{@{}llrr@{}}
\toprule
\shortstack{Capacity}
& Cell mode (P/E)
& \shortstack{1$\times$ lifetime}
& \shortstack{5$\times$ lifetime}\\
\midrule
512 GB & TLC (3k)           & 3.0 yrs        & 3.2 yrs \\
375 GB & MLC (10k) & 7.3 yrs & 7.9 yrs \\
170 GB & pSLC (100k)        & 33.2 yrs      & 30.6 yrs\\
\bottomrule
\end{tabular}
\end{table}

\paragraph{Sensitivity to HBF capacity.}

We sweep HBF capacity per stack over 512, 375, and 170 GB. We additionally evaluate a more conservative device model in which 512~GB, 375~GB, 170~GB stack uses TLC, MLC, pSLC, respectively.
Table~\ref{tab:sensitivity} shows increasing capacity to 512 GB does not improve
completion time, while reducing it to 170 GB affects only the
5$\times$ trace (+5.83\%) and changes energy savings by less than 0.2
percentage points. As Table~\ref{tab:capacity-lifetime} shows, TLC HBF wears
out within about 3 years, whereas pSLC at one-third the capacity lasts over
30 years. Large HBF capacity yields diminishing returns, and trading it for endurance costs little performance. MLC provides a 7.3-year lifetime for multi-GPU GLM-5.2 serving. Single-GPU Qwen3-32B serving writes more frequently due to its smaller HBF capacity and lower communication overhead, lasting only 1.36 years under MLC, which makes pSLC necessary.


\section{Conclusion}
For LLM serving, an HBF system with HBM–HBF–host hierarchy and buffered cache-aware scheduling reduces completion time by 21.1-86.9\% and modeled energy by up to 62.4\%, while keeping HBF write lifetime practical. Our results demonstrate that HBF is most effective as an actively utilized lower tier in the memory hierarchy, with careful data placement and scheduling essential to achieving its throughput and energy-efficiency benefits.


\section*{Acknowledgements}

We acknowledge gracious support from the FuriosaAI, Apple, NVIDIA, Macronix, Citadel, Meta, and Panasonic.
Furthermore, we appreciate support from Google Cloud, the Google TRC team, and specifically Jonathan Caton, Divvy Thakkar, and Prof. David Patterson.
Prof. Keutzer's lab is also sponsored by funding through BDD, BAIR, and Humanoid robotics.
We also acknowledge support by the Director, Office of Science, Office of Advanced Scientific Computing Research, of the U.S. Department of Energy under Contract No. DE-AC02-05CH11231.
Amir Gholaminejad acknowledges DOE SciGPT grant and UCOP.
Michael W. Mahoney acknowledges DARPA, NSF, the DOE Competitive Portfolios grant, and
the DOE SciGPT grant. Our conclusions do not necessarily
reflect the position or the policy of our sponsors, and no
official endorsement should be inferred.

\clearpage
\bibliographystyle{IEEEtran}
\bibliography{references}

\begin{thebibliography}{10}
\providecommand{\url}[1]{#1}
\csname url@samestyle\endcsname
\providecommand{\newblock}{\relax}
\providecommand{\bibinfo}[2]{#2}
\providecommand{\BIBentrySTDinterwordspacing}{\spaceskip=0pt\relax}
\providecommand{\BIBentryALTinterwordstretchfactor}{4}
\providecommand{\BIBentryALTinterwordspacing}{\spaceskip=\fontdimen2\font plus
\BIBentryALTinterwordstretchfactor\fontdimen3\font minus
  \fontdimen4\font\relax}
\providecommand{\BIBforeignlanguage}[2]{{%
\expandafter\ifx\csname l@#1\endcsname\relax
\typeout{** WARNING: IEEEtran.bst: No hyphenation pattern has been}%
\typeout{** loaded for the language `#1'. Using the pattern for}%
\typeout{** the default language instead.}%
\else
\language=\csname l@#1\endcsname
\fi
#2}}
\providecommand{\BIBdecl}{\relax}
\BIBdecl

\bibitem{gholami2024ai}
A.~Gholami, Z.~Yao, S.~Kim, C.~Hooper, M.~W. Mahoney, and K.~Keutzer, ``Ai and
  memory wall,'' \emph{IEEE Micro}, vol.~44, no.~3, pp. 33--39, 2024.

\bibitem{flexgen2023}
Y.~Sheng, L.~Zheng, B.~Yuan, Z.~Li, M.~Ryabinin, B.~Chen, P.~Liang, C.~R{\'e},
  I.~Stoica, and C.~Zhang, ``Flexgen: High-throughput generative inference of
  large language models with a single gpu,'' in \emph{International Conference
  on Machine Learning}.\hskip 1em plus 0.5em minus 0.4em\relax PMLR, 2023, pp.
  31\,094--31\,116.

\bibitem{vllm2023}
W.~Kwon, Z.~Li, S.~Zhuang, Y.~Sheng, L.~Zheng, C.~H. Yu, J.~Gonzalez, H.~Zhang,
  and I.~Stoica, ``Efficient memory management for large language model serving
  with pagedattention,'' in \emph{Proceedings of the 29th symposium on
  operating systems principles}, 2023, pp. 611--626.

\bibitem{sglang2024}
L.~Zheng, L.~Yin, Z.~Xie, C.~Sun, J.~Huang, C.~H. Yu, S.~Cao, C.~Kozyrakis,
  I.~Stoica, J.~E. Gonzalez \emph{et~al.}, ``Sglang: Efficient execution of
  structured language model programs,'' \emph{Advances in neural information
  processing systems}, vol.~37, pp. 62\,557--62\,583, 2024.

\bibitem{autellix2025}
\BIBentryALTinterwordspacing
M.~Luo, X.~Shi, C.~Cai, T.~Zhang, J.~Wong, Y.~Wang, C.~Wang, Y.~Huang, Z.~Chen,
  J.~E. Gonzalez, and I.~Stoica, ``Agentix: An efficient serving engine for
  {LLM} agents as general programs,'' in \emph{23rd USENIX Symposium on
  Networked Systems Design and Implementation (NSDI 26)}.\hskip 1em plus 0.5em
  minus 0.4em\relax Renton, WA: USENIX Association, May 2026, pp. 2443--2459.
  [Online]. Available:
  \url{https://www.usenix.org/conference/nsdi26/presentation/luo}
\BIBentrySTDinterwordspacing

\bibitem{continuum2025}
H.~Li, R.~He, Q.~Mang, Q.~Zhang, H.~Mao, X.~Chen, H.~Zhou, H.~Zhang, A.~Cheung,
  J.~Gonzalez \emph{et~al.}, ``Continuum: Efficient and robust multi-turn llm
  agent scheduling with kv cache time-to-live,'' \emph{arXiv preprint
  arXiv:2511.02230}, 2025.

\bibitem{sandiskhbf2025}
\BIBentryALTinterwordspacing
{Sandisk}, ``{Sandisk Unveils the Future of Memory Architecture for {AI}:
  Introducing High Bandwidth Flash},'' Tech brief / fact sheet, Jul. 2025.
  [Online]. Available:
  \url{https://documents.sandisk.com/content/dam/asset-library/en_us/assets/public/sandisk/collateral/company/Sandisk-HBF-Fact-Sheet.pdf}
\BIBentrySTDinterwordspacing

\bibitem{h32026}
M.~Ha, E.~Kim, and H.~Kim, ``H3: Hybrid architecture using high bandwidth
  memory and high bandwidth flash for cost-efficient llm inference,''
  \emph{IEEE Computer Architecture Letters}, vol.~25, no.~1, pp. 49--52, 2026.

\bibitem{mapatterson2026}
X.~Ma and D.~Patterson, ``Challenges and research directions for large language
  model inference hardware,'' \emph{Computer}, vol.~59, no.~5, pp. 55--64,
  2026.

\bibitem{orca2022}
G.-I. Yu, J.~S. Jeong, G.-W. Kim, S.~Kim, and B.-G. Chun, ``Orca: A distributed
  serving system for {Transformer-Based} generative models,'' in \emph{16th
  USENIX Symposium on Operating Systems Design and Implementation (OSDI
  22)}.\hskip 1em plus 0.5em minus 0.4em\relax Carlsbad, CA: USENIX
  Association, Jul. 2022, pp. 521--538.

\bibitem{xie2026hisparse}
Z.~Xie, Z.~Huang, T.~Huang, Z.~Xu, R.~Ma, and C.~Kozyrakis, ``Hisparse: Scaling
  sparse-attention decoding with hierarchical kv cache management,''
  \emph{arXiv preprint arXiv:2608.07009}, 2026.

\bibitem{lmcache2025}
Y.~Liu, Y.~Cheng, J.~Yao, Y.~An, X.~Chen, S.~Feng, Y.~Huang, S.~Shen, R.~Zhang,
  K.~Du \emph{et~al.}, ``Lmcache: An efficient kv cache layer for
  enterprise-scale llm inference,'' \emph{arXiv preprint arXiv:2510.09665},
  2025.

\bibitem{mooncake2025}
R.~Qin, Z.~Li, W.~He, J.~Cui, H.~Tang, F.~Ren, T.~Ma, S.~Cai, Y.~Zhang,
  M.~Zhang \emph{et~al.}, ``Mooncake: A kvcache-centric disaggregated
  architecture for llm serving,'' \emph{ACM Transactions on Storage}, vol.~22,
  no.~4, pp. 1--38, 2026.

\bibitem{llminaflash2024}
K.~Alizadeh, S.~I. Mirzadeh, D.~Belenko, S.~Khatamifard, M.~Cho, C.~C.
  Del~Mundo, M.~Rastegari, and M.~Farajtabar, ``Llm in a flash: Efficient large
  language model inference with limited memory,'' in \emph{Proceedings of the
  62nd Annual Meeting of the Association for Computational Linguistics (Volume
  1: Long Papers)}, 2024, pp. 12\,562--12\,584.

\bibitem{hifc2025}
\BIBentryALTinterwordspacing
I.~Jeong, S.~Woo, S.~Namkung, and D.~Jeon, ``Hi{FC}: High-efficiency
  flash-based {KV} cache swapping for scaling {LLM} inference,'' in \emph{The
  Thirty-ninth Annual Conference on Neural Information Processing Systems},
  2025. [Online]. Available: \url{https://openreview.net/forum?id=onhjdWCxZY}
\BIBentrySTDinterwordspacing

\bibitem{zhao2026llm}
\BIBentryALTinterwordspacing
S.~Zhao, M.~Kim, C.~R.~C. Hooper, L.~Manolache, M.~W. Mahoney, S.~Shao,
  K.~Keutzer, and A.~Gholami, ``{LLM} inference in a flash!'' in \emph{Machine
  Learning for Computer Architecture and Systems 2026}, 2026. [Online].
  Available: \url{https://openreview.net/forum?id=gSphYexssc}
\BIBentrySTDinterwordspacing

\bibitem{hbfsucks2026}
Z.~Li, Z.~Bian, X.~Huang, Y.~Zhao, G.~Sun, and Y.~Zhuo, ``Hbf sucks? a
  full-stack characterization of high-bandwidth flash for kv-centric llm
  serving,'' \emph{arXiv preprint arXiv:2608.11668}, 2026.

\bibitem{petrucci2026hbf}
\BIBentryALTinterwordspacing
V.~Petrucci, F.~Zacarias, and V.~Tanna, ``{Is High-Bandwidth Flash All You
  Need?}'' in \emph{3rd Workshop on Hot Topics in System Infrastructure
  (HotInfra)}, Raleigh, NC, USA, 2026, workshop co-located with ISCA 2026, June
  28, 2026. [Online]. Available:
  \url{https://hotinfra.org/2026/papers/hotinfra26-final83.pdf}
\BIBentrySTDinterwordspacing

\bibitem{son2026hbf}
D.~Son, Y.~Park, H.~Cho, H.~Ham, O.~Mutlu, S.~Lee, G.~Kim, and J.~Park,
  ``Exploring high-bandwidth flash for modern llm inference: Opportunities and
  challenges,'' \emph{IEEE Computer Architecture Letters}, 2026.

\bibitem{llmservingsim2026}
J.~Cho, H.~Choi, G.~Heo, and J.~Park, ``Llmservingsim 2.0: A unified simulator
  for heterogeneous and disaggregated llm serving infrastructure,'' in
  \emph{2026 IEEE International Symposium on Performance Analysis of Systems
  and Software (ISPASS)}.\hskip 1em plus 0.5em minus 0.4em\relax IEEE, 2026,
  pp. 1--14.

\bibitem{memexplorer2026}
H.~Wu, Z.~Cao, Y.~Lai, B.~Lou, J.~Nie, C.~Xiao, T.~Adeniran, K.~Lau, P.~Forys,
  K.~Johar \emph{et~al.}, ``Memexplorer: Navigating the heterogeneous memory
  design space for agentic inference npus,'' \emph{arXiv preprint
  arXiv:2604.16007}, 2026.

\bibitem{agrawal2024sarathiserve}
A.~Agrawal, N.~Kedia, A.~Panwar, J.~Mohan, N.~Kwatra, B.~Gulavani, A.~Tumanov,
  and R.~Ramjee, ``Taming $\{$Throughput-Latency$\}$ tradeoff in $\{$LLM$\}$
  inference with $\{$Sarathi-Serve$\}$,'' in \emph{18th USENIX symposium on
  operating systems design and implementation (OSDI 24)}, 2024, pp. 117--134.

\bibitem{liu2025deepseek}
A.~Liu, A.~Mei, B.~Lin, B.~Xue, B.~Wang, B.~Xu, B.~Wu, B.~Zhang, C.~Lin,
  C.~Dong \emph{et~al.}, ``Deepseek-v3. 2: Pushing the frontier of open large
  language models,'' \emph{arXiv preprint arXiv:2512.02556}, 2025.

\bibitem{megatron2019}
M.~Shoeybi, M.~Patwary, R.~Puri, P.~LeGresley, J.~Casper, and B.~Catanzaro,
  ``Megatron-lm: Training multi-billion parameter language models using model
  parallelism,'' \emph{arXiv preprint arXiv:1909.08053}, 2019.

\bibitem{gshard2021}
\BIBentryALTinterwordspacing
D.~Lepikhin, H.~Lee, Y.~Xu, D.~Chen, O.~Firat, Y.~Huang, M.~Krikun, N.~Shazeer,
  and Z.~Chen, ``{\{}GS{\}}hard: Scaling giant models with conditional
  computation and automatic sharding,'' in \emph{International Conference on
  Learning Representations}, 2021. [Online]. Available:
  \url{https://openreview.net/forum?id=qrwe7XHTmYb}
\BIBentrySTDinterwordspacing

\bibitem{deepseekv32024}
A.~Liu, B.~Feng, B.~Xue, B.~Wang, B.~Wu, C.~Lu, C.~Zhao, C.~Deng, C.~Zhang,
  C.~Ruan \emph{et~al.}, ``Deepseek-v3 technical report,'' \emph{arXiv preprint
  arXiv:2412.19437}, 2024.

\bibitem{deepseek_lplb_2025}
{DeepSeek-AI}, ``Lplb: Linear-programming-based load balancer for
  mixture-of-experts models,'' \url{https://github.com/deepseek-ai/LPLB}, 2025.

\bibitem{indexcache2026}
Y.~Bai, Q.~Dong, T.~Jiang, X.~Lv, Z.~Du, A.~Zeng, J.~Tang, and J.~Li,
  ``Indexcache: Accelerating sparse attention via cross-layer index reuse,''
  \emph{arXiv preprint arXiv:2603.12201}, 2026.

\bibitem{qwen32025}
A.~Yang, A.~Li, B.~Yang, B.~Zhang, B.~Hui, B.~Zheng, B.~Yu, C.~Gao, C.~Huang,
  C.~Lv \emph{et~al.}, ``Qwen3 technical report,'' \emph{arXiv preprint
  arXiv:2505.09388}, 2025.

\bibitem{lmcache-agentic-dataset-2026}
\BIBentryALTinterwordspacing
{LMCache}, ``{LMCache Agentic Dataset: Multi-Turn LLM Agent Sessions for KV
  Cache Benchmarking},'' 2026. [Online]. Available:
  \url{https://huggingface.co/datasets/sammshen/lmcache-agentic-traces}
\BIBentrySTDinterwordspacing

\bibitem{micronhbm3e}
\BIBentryALTinterwordspacing
{Micron Technology}, ``{HBM3E}.'' [Online]. Available:
  \url{https://www.micron.com/products/memory/hbm/hbm3e}
\BIBentrySTDinterwordspacing

\bibitem{lenovob200}
\BIBentryALTinterwordspacing
{Lenovo}, ``{ThinkSystem NVIDIA HGX B200 180GB 1000W GPU Product Guide},''
  Lenovo Press. [Online]. Available:
  \url{https://lenovopress.lenovo.com/lp2226-thinksystem-nvidia-b200-180gb-1000w-gpu}
\BIBentrySTDinterwordspacing

\bibitem{moon2023packaging}
K.-I. Moon, H.-Y. Son, and K.~Lee, ``Advanced packaging technologies in memory
  applications for future generative ai era,'' in \emph{2023 International
  Electron Devices Meeting (IEDM)}, 2023, pp. 1--4.

\bibitem{tsmcn52020}
\BIBentryALTinterwordspacing
{TSMC}, ``{TSMC} showcases leading technologies at online technology symposium
  and {OIP} ecosystem forum,'' 2020. [Online]. Available:
  \url{https://pr.tsmc.com/english/news/2729}
\BIBentrySTDinterwordspacing

\bibitem{tsmcn4p2021}
\BIBentryALTinterwordspacing
------, ``{TSMC} expands advanced technology leadership with {N4P} process,''
  2021. [Online]. Available: \url{https://pr.tsmc.com/english/news/2874}
\BIBentrySTDinterwordspacing

\bibitem{glm2026}
A.~Zeng, X.~Lv, Z.~Hou, Z.~Du, Q.~Zheng, B.~Chen, D.~Yin, C.~Ge, C.~Huang,
  C.~Xie \emph{et~al.}, ``Glm-5: from vibe coding to agentic engineering,''
  \emph{arXiv preprint arXiv:2602.15763}, 2026.

\bibitem{samsung2024vnand}
\BIBentryALTinterwordspacing
{Samsung Electronics}, ``{Samsung V-NAND: A Landmark of the Hyperscale AI
  Era},'' Jul. 2024. [Online]. Available:
  \url{https://news.samsungsemiconductor.com/global/samsung-v-nand-a-landmark-of-the-hyperscale-ai-era/}
\BIBentrySTDinterwordspacing

\bibitem{pan2026measuring}
\BIBentryALTinterwordspacing
M.~Pan, N.~Arabzadeh, R.~Cogo, Y.~Zhu, A.~Xiong, L.~A. Agrawal, H.~Mao,
  E.~Shen, S.~Pallerla, L.~Patel, S.~Liu, T.~Shi, X.~Liu, J.~Q. Davis,
  E.~Lacavalla, A.~Basile, S.~Yang, P.~Castro, D.~Kang, K.~Sen, D.~Song, J.~E.
  Gonzalez, I.~Stoica, M.~Zaharia, and M.~Ellis, ``Measuring agents in
  production,'' in \emph{Forty-third International Conference on Machine
  Learning}, 2026. [Online]. Available:
  \url{https://openreview.net/forum?id=mWxEAgz3xu}
\BIBentrySTDinterwordspacing

\end{thebibliography}
\end{document}